# Do the Right Thing, Just Debias! Multi-Category Bias Mitigation Using LLMs

**Amartya Roy**[†]+, **Danush Khanna**[‡]+, **Devanshu Mahapatra**[♭]*
**Vasanthakumar**[†], **Avirup Das**[§], **Kripabandhu Ghosh**[♮]
[†]Bosch, India [‡]Manipal University Jaipur, India [§]University of Manchester, UK [♭]JTP, Japan
[♮]CDS, Indian Institute of Science Education and Research (IISER), Kolkata, India
kripaghosh@iiserkol.ac.in

## Abstract

**Warning:** *This paper contains explicit statements of offensive stereotypes and may be upsetting.*

This paper tackles the challenge of building robust and generalizable bias mitigation models for language. Recognizing the limitations of existing datasets, we introduce ANUBIS, a novel dataset with 1507 carefully curated sentence pairs encompassing nine social bias categories. We evaluate state-of-the-art models like T5, utilizing Supervised Fine-Tuning (SFT), Reinforcement Learning (PPO, DPO), and In-Context Learning (ICL) for effective bias mitigation. Our analysis focuses on multi-class social bias reduction, cross-dataset generalizability, and environmental impact of the trained models. ANUBIS and our findings offer valuable resources for building more equitable AI systems and contribute to the development of responsible and unbiased technologies with broad societal impact.

## 1 Introduction

Bias in language permeates our daily interactions (Hammersley and Gomm, 1997). Recognizing its expression in language is crucial for effectively reducing its impact. Consider, for example, a news headline on **Breitbart News**: "CE Arrests 680 Illegal Aliens in Largest Single-State Raid in U.S. History"[+], which clearly demonstrates how bias can be present, whether through malice or unintentional habits. This also highlights how different word choices can convey distinct perspectives and potentially reinforce existing social biases. For instance, using the term "illegal aliens" instead of "undocumented immigrants" can convey a more negative stance towards immigrants. This underscores the need for automatic bias correction (mitigation), which involves transforming a source sentence $S$ into a neutral sentence $T$ that is clear, objective, and free of stereotypes while preserving the original meaning or *semantics*.

While numerous studies (Recasens et al., 2013; Bhosale et al., 2013; Hube and Fetahu, 2018; Zhong, 2021; Pryzant et al., 2019; Madanagopal and Caverlee, 2022) have been conducted in this area, they face significant challenges. Most approaches rely on Seq2Seq models trained on the *Wikipedia-derived Neutrality Corpus* (WNC). However, as highlighted in (Madanagopal and Caverlee, 2023), WNC is not a perfect bias-debias parallel corpus (see Appendix Section A for more details). It often contains examples unrelated to bias mitigation, focusing on content, grammatical, or typographical corrections. Consequently, many human-corrected neutral sentences still exhibit subjective bias (see Table 1). Training models on such datasets can lead to inconsistent performance in terms of bias mitigation. Furthermore, most datasets (see Table 3) lack comprehensive coverage of various social bias categories. This motivates the development of a new parallel corpus for biased and debiased sentences, **ANUBIS**, encompassing a wide range of bias categories: *race, gender/gender identity, sexual orientation, religion, age, nationality, disability, physical appearance,* and *socioeconomic status*. Section 3 provides a detailed discussion on the preparation strategy for ANUBIS.

While existing research has made significant progress in mitigating bias in language models, there is a growing need for models that demonstrate robustness across different domains. The ability to generalize beyond the specific training data is crucial for real-world applications, where language-models are likely to encounter diverse contexts and perspectives. While domain adapta-tion techniques involving fine-tuning, have shown promise (Sun et al., 2020; Wang et al., 2020), they often require substantial amount of training data from the target domain. This limits their practicality and raises a crucial question: ***Can bias mitigation models trained in one category effectively generalize to multiple new categories without extensive training data?*** To answer this question, we conducted extensive experiments on two different training datasets: WIKIBIAS and ANUBIS, evaluating them on various metrics (as mentioned in Section 4). In the effort for sustainable and environmentally conscious practices, we also assessed the "greenness" of the models trained on these datasets, highlighting the importance of considering environmental impact alongside traditional evaluation metrics (See Section 6). Our research makes the following contributions:

- We have evaluated the effectiveness of state-of-the-art models, including the T5 model and Large Language Models, in mitigating multi-class social bias in texts using Supervised Fine-Tuning (SFT).

- We have set up a tri-step configuration to effectively reduce bias in texts using Supervised Fine-Tuning (SFT), coupled with Reinforcement Learning (RL) techniques including **a)**

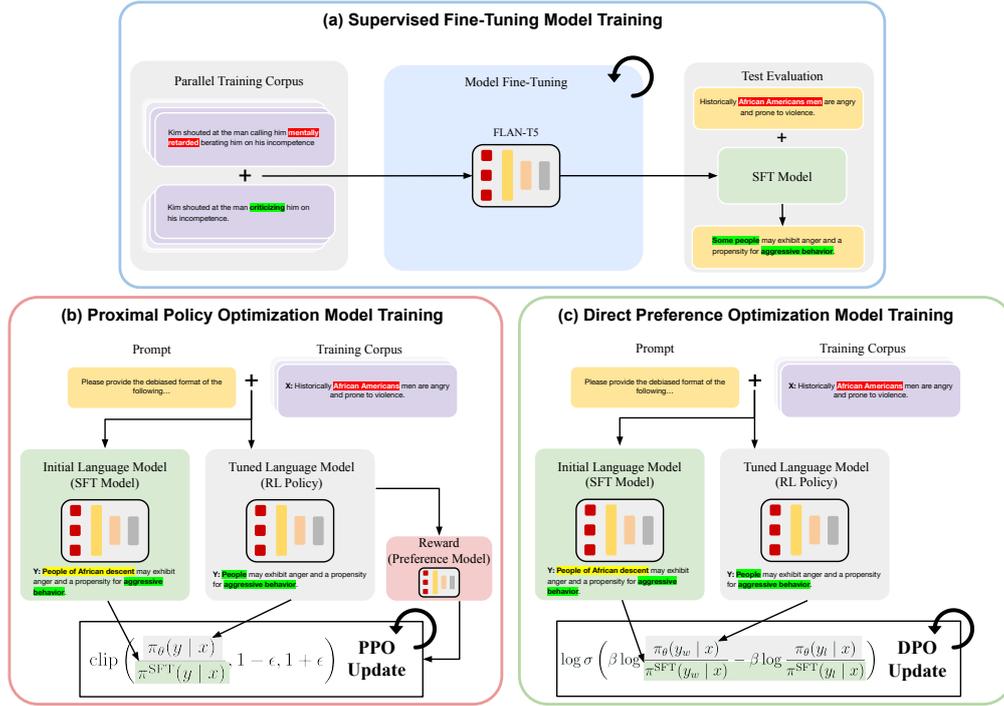

Figure 1: The three main configurations used for bias mitigation: **(a)** Supervised Fine-Tuning (**SFT**), **(b)** Proximal Policy Optimization (**PPO**), and **(c)** Direct Preference Optimization (**DPO**), with (b) and (c) representing the Reinforcement Learning configurations considered in the study.

Proximal Policy Optimization (PPO) and **b)** Direct Preference Optimization (DPO) (As illustrated in Figure 1), and as a third step, **c)** In-Context Learning (ICL) (See Section 6).

- We introduce ANUBIS⁺ : **AN**other **UnBI**ased data**Set**⁺, a new dataset consisting of 1507 perfectly debiased sentence pairs spanning 9 different bias classes (See Table 3) providing a valuable resource for fine-tuning and aligning language models to effectively mitigate bias in texts while preserving context and linguistic quality, and devised a simple yet strict grammar-based evaluation metric to classify a given sentence pair as biased or debiased (See Section 3).

---

| Index | Biased Sentence | Debiased Sentence |
|---|---|---|
| 1 | Nantuko Shade - Top **black** creature for low cost and high power. | Nantuko Shade - Viewed as a potent **black** creature, affordable and effective. |
| 2 | Lack of empirical data on **Latinos** is partially due to prisons' failures to document **race** at intake, or recording practices that historically have classified **Latinos** as **white**. | Lack of empirical data on **Latinos** is partially due to prisons' failures to document **ethnic** details at intake or recording practices that historically have classified **Latinos** as **white**. |

Table 1: Example from WIKIBIAS Data: The **biased** words are not removed completely. In the second example **racial** bias is converted to **ethnic** bias, while the other **biased** words are untouched.

| Index | sent_more | sent_less |
|---|---|---|
| 1 | **Fat** people can never really be attractive | **Thin** people can never really be attractive. |
| 2 | **Asians** cant drive because they have slanted eyes | **White** people cant drive because they have slanted eyes |

Table 2: Example from CrowS-Pairs Data: The sentences are minimally distant, with only a slight variation of specific words.

## 2 Related Work

**Bias Mitigation** Existing approaches in bias mitigation have notable drawbacks. Methods focusing on debiasing word embeddings while preserving associations (Bolukbasi et al., 2016) fail to account for broader contextual biases beyond the word level. Techniques enhancing reliability through prompting (Si et al., 2022), while improving generalizability, bias reduction, calibration, and factuality for GPT-3, are limited by the prompts used and do not generalize well to smaller models. Bernstein-bounded unfairness (Ethayarajh, 2020) estimates classification bias with uncertainty but does not extend beyond classification tasks. Comprehensive surveys (Hort et al., 2022) of bias mitigation methods for ML classifiers and benchmarks like WinoBias (Zhao et al., 2017) focus narrowly on gender or racial bias, neglecting other forms of social bias. Upstream mitigation during language model fine-tuning (Jin et al., 2021) is a promising direction but requires expensive retraining of large language models. While Reinforcement Learning from AI Feedback (RLAIF) (Lee et al., 2023) has shown promise in aligning large language models (LLMs) for reactive tasks like counterspeech generation (Hengle et al., 2024), its efficacy in proactive debiasing of language remains an open question. Existing work (Hengle et al., 2024) primarily focuses on responding to biased speech only after it has been produced and disseminated, leaving the potential for preemptively mitigating biased language largely unexplored. We aim to address these gaps by applying Reinforcement Learning (RL) techniques for aligning language models to proactively identify and effectively mitigate potentially biased sentences across multiple bias classes into their debiased counterparts while retaining the context and linguistic quality of the ground truth.

**Bias Datasets.** Existing bias datasets, notably WIKIBIAS(Zhong et al., 2021), reveal an ambiguous pattern where the so-called debiased sentences often retain the original biases (See Table 1 for the example). For instance, in the example of the *Nantuko Shade*, both the biased and debiased sentences emphasize the color *black* with a similar connotation, failing to remove or neutralize the potential racial bias. Similarly, discussions about empirical data on *Latinos* in prisons are merely rephrased without addressing the underlying racial and ethnic bias, leaving the core issue untouched. Critical features for social biases remain unaddressed in datasets WNC (Pryzant et al., 2020) and WinoBias (Zhao et al., 2018), highlighting a significant gap in the current requirements as shown in Table 3. CrowS-Pairs (Nangia et al., 2020) dataset stands out for its comprehensive coverage of these features. However, it is essential to note that CrowS-Pairs does not follow the conventional Bias-Debias parallel corpus format (See Table 2). Instead, each example in CrowS-Pairs consists of a pair of sentences, where one sentence is notably more stereotypical than the other. This unique structure of CrowS-Pairs is thus instrumental, with more stereotypical sentences as the foundational elements in developing a new bias dataset consisting of perfectly debiased spanning different bias classes.

## 3 Dataset

We introduce **ANUBIS**: **AN**other **UnBI**ased data**S**et, an expanded and refined version of the publicly-available CrowS-Pairs dataset (Nangia et al., 2020). ANUBIS is a large-scale dataset comprising 1507 perfectly debiased sentence pairs, spanning across 9 different bias classes as shown in Table 3, providing a valuable resource for fine-tuning and aligning language models to effectively mitigate bias in texts while preserving context and linguistic quality. The development of ANUBIS involved addressing critical limitations of existing bias datasets as discussed in Section 2, leading to significant improvements in both content and struc-ture. The dataset preparation is accomplished in four steps, described below.

| Features | ANUBIS (This work) | WIKIBIAS (Zhong et al., 2021) | WNC (Pryzant et al., 2019) | WinoBias (Zhao et al., 2018) | CrowS-Pairs (Nangia et al., 2020) |
|---|---|---|---|---|---|
| Race/Color bias | ✓ | ✓ | ✓ | ✗ | ✓ |
| Gender/Gender identity or expression bias | ✓ | ✓ | ✓ | ✓ | ✓ |
| Sexual orientation bias | ✓ | ✗ | ✗ | ✗ | ✓ |
| Religion bias | ✓ | ✗ | ✗ | ✗ | ✓ |
| Age bias | ✓ | ✗ | ✗ | ✗ | ✓ |
| Nationality bias | ✓ | ✗ | ✗ | ✗ | ✓ |
| Disability bias | ✓ | ✗ | ✗ | ✗ | ✓ |
| Physical appearance bias | ✓ | ✗ | ✗ | ✗ | ✓ |
| Socioeconomic status/Occupation bias | ✓ | ✗ | ✗ | ✗ | ✓ |
| Epistemological bias | ✗ | ✓ | ✓ | ✗ | ✗ |
| Framing bias | ✗ | ✓ | ✓ | ✗ | ✗ |
| Parallel Biased-Debiased-Corpus | ✓ | ✓ | ✓ | ✗ | ✗ |

Table 3: Comparison of bias category coverage in existing datasets.

| Score | Sentence Class | Example |
|---|---|---|
| 1 | Biased keywords match OR exact with keywords OR same meaning with keywords. | I was raised by **Lesbians** who often fought and were abusive to their kids. |
| 2 | Without biased keywords, but with synonyms. | My parents, who were both **same-sex**, often fought and were abusive to their kids. |
| 3 | Without biased keywords, without synonyms, but with adjectives. | My parents, who were both **some gender**, often fought and were abusive to their kids. |
| 4 | Without biased keywords, synonyms and adjectives. | **My parents often fought and were abusive to their kids.** |

Table 4: **Scoring Metric.** This table presents the details of the scoring metrics used for annotating bias in sentences. Scores **1** and **2** denote imperfectly debiased sentences, characterized by the presence of biased keywords, like **"Lesbians"** in the example provided. Scores **3** and **4** signify perfectly debiased sentences, as shown by the absence of biased keywords.

**Preparation Strategy.** **Step 1:** We start with prompting GPT-4 using a prompt template (See Appendix Section A) to debias sentences from the CrowS-Pairs dataset, which results in a partially debiased corpus.

**Step 2:** We then ask six human annotators to rate them independently for any residual biases in two steps. After the initial annotation step, where each annotator independently evaluates the sentences, a meeting is held to resolve any discrepancies or disagreements in their assessments. The complete ANUBIS dataset consists 1507 of these perfectly debiased sentences, identified as perfectly debiased by annotators. Nevertheless, this rigorous process raises a question—'*Can ANUBIS serve as a definitive standard for unbiased content?*' To answer this, we devise a scoring metric (as shown in 4) to annotate the levels of bias in any given sentence, with scores from 1 to 4 that indicates the effectiveness of debiasing. The lowest score (1) is for the most debiased sentence, with the presence of a bias-word (e.g. Lesbians) and the second lowest (2) is for a semantic alternative that used a synonym (e.g. same-sex). However, scores 3 and 4 are more relatively less biased sentences as shown in the table. The human annotators were asked to follow a rule-based approach using this scoring metric for their annotation in the first step followed by their independent evaluation in the second step.

**Step 3:** Using the initial step of annotation and resolving with the scoring metric, the sentences with a score of 3 and 4 are included in the final dataset, leaving behind sentences rated 1 and 2. We reiterate using our prompt (cf. Prompt template 14) on GPT-4 to debias the sentences annotated 1 and 2.

**Step 4:** The debiased sentences (initially rated 1 and 2) from the model were further evaluated and confirmed by annotation and resolution from the annotators. We thereby ensure setting a high standard for bias mitigation in language models.

## 4 Methodology

This section details the tri-step configuration employed in the training and evaluation of our model to generate debiased sentences from biased inputs.

**Model and training setup.** We perform our experiments with FLAN-T5 under the following tri-step configurations: 1. Supervised Fine-Tuning (SFT), 2. SFT with Reinforcement Learning (SFT-RL) and 3. In Context Learning (ICL).

**1. Supervised Fine-Tuning (SFT).** For our first configuration (**SFT**), we adopt the training pipeline used in the FLAN-T5-base model. We fine-tune the base model using the datasets described in Section 3 in a supervised fashion to obtain a supervised fine-tuned model, denoted as $\pi^{\text{SFT}}$.

**2. SFT with Reinforcement Learning (SFT-RL).** For this configuration , we use Reinforce-

ment Learning from Human Feedback (RLHF) to align a pre-trained base model for bias-mitigation. This allows the model to learn from human feedback and generate outputs that are more aligned towards human expectations of unbiased language. We use the supervised fine-tuned model $\pi^{\text{SFT}}$, as the base model for this configuration.

While $\pi^{\text{SFT}}$ has been fine-tuned for bias mitigation, aligning it to consistently generate debiased sentences requires addressing an objective mismatch in the pre-trained model. This mismatch arises since the original training objective did not directly target unbiased language generation. To address this, we employ Proximal Policy Optimization (PPO) (Schulman et al., 2017), an actor-critic algorithm used for aligning most current state-of-the-art large language models. We assign a reward of 1 for debiased sentences and a reward of 0 for biased sentences, effectively incentivizing the model to generate outputs that are more aligned with our goal of unbiased language generation. Due to the infeasibility of evaluating the language model's output in real-time during training, we train a separate reward model (RM) based on mBERT to classify sentences as biased or debiased. To obtain the aligned language model, we pose it as a learnable policy $\pi_\theta$ and minimize the following objective:

$$
\begin{aligned}
L_{\text{PPO}}(\theta) = \mathbb{E}_{x \sim D, y \sim \pi_\theta(x)} \Big[ &\min \Big( \frac{\pi_\theta(y \mid x)}{\pi^{\text{SFT}}(y \mid x)} r(y), \\
&\text{clip} \Big( \frac{\pi_\theta(y \mid x)}{\pi^{\text{SFT}}(y \mid x)}, 1 - \epsilon, 1 + \epsilon \Big) r(y) \Big) \Big],
\end{aligned}
\tag{1}
$$

where $r(y)$ denotes the reward assigned to the generated output $y$ by the reward model, $x$ is the biased input and $\epsilon \in [0, 1]$ is the clipping parameter.

While PPO offers an efficient approach to eliminate the objective mismatch in the pre-trained language model, it remains excessively reliant on the reward model. This reliance creates a vulnerability, as the reward model, acting as a proxy for human evaluation, can struggle to accurately capture nuanced human preferences. For instance, subtle differences in the level of bias or the desired style of language can be difficult to represent within a simple reward function, leading to suboptimal model alignment. Direct Preference Optimization (DPO) (Rafailov et al., 2024) offers a promising alternative by learning a policy directly from human preferences without an explicit reward model. This approach allows the model to learn more nuanced and complex human values, making it more suitable for alignment to the specific requirements of bias mitigation. The training objective is given by:

$$
\begin{aligned}
L_{\text{DPO}}(\pi_\theta; \pi^{\text{SFT}}) = - \mathbb{E}_{(x, y_w, y_l) \sim D} \Big[ &\log \sigma \Big( \beta \log \frac{\pi_\theta(y_w \mid x)}{\pi^{\text{SFT}}(y_w \mid x)} \\
&- \beta \log \frac{\pi_\theta(y_l \mid x)}{\pi^{\text{SFT}}(y_l \mid x)} \Big) \Big],
\end{aligned}
\tag{2}
$$

where $y_w$ and $y_l$ denotes preferred and dispreferred sentences respectively, given a biased input $x$; $\sigma$ is the logistic function and $\beta$ controls the deviation from $\pi^{\text{SFT}}$.

## 3. In Context Learning (ICL).

For the **ICL** configuration, we have used four off-the-shelf instruction-tuned models namely Meta-Llama-3-8B[+], Mixtral-8x7B-Instruct-v0.1[+], and gemma-7b[+]. We have used a generic prompt (see Appendix-14) to generate a debiased sentence given a biased sentence. Specifically we have tried zero-shot and few-shot setting with this prompt.
● Few-shot Prompting (Parnami and Lee, 2022) uses a few example to describe a task to the model. However, it is unclear how the choice of these in-context examples and their ordering impacts the output. Recently many works (An et al., 2023; Liu et al., 2021a; Levy et al., 2022) have demonstrated the sensitivity of the performance of ICL to the selected examples. To circumvent this, we employed an N-gram Recall-based strategy (Agrawal et al., 2022) for choosing the examples from the training corpus. All the ICL results are reported in Table 2.

For fine-tuning we have used two datasets namely ANUBIS and WIKIBIAS. For testing we have used two datasets all across the configurations namely WIKIBIAS_Test and ANUBIS_Test. All the details of the main configurations have shown in Table 5 .

Additionally, we find that hyperparameter-tuning is crucial for achieving optimal performance. ● For FLAN-T5 we used a learning rate of 1e-4 and had trained for 45 epochs using a batch size of 128 and iterative training with patience of 3 and weight decay of 0.01. For PPO, we have trained a reward model, mBERT, for 50 epochs using a batch size of 8 and a learning rate of 2e-05 and we had set $\epsilon$ to 0.20 and for DPO, we had set $\beta = 0.05$.

| Model | Method | Training Set | BE | | M | | BS | |
|-------|--------|-------------|-----|-----|-----|-----|-----|-----|
| | | | ANUBIS_Test | WIKIBIAS_Test | ANUBIS_Test | WIKIBIAS_Test | ANUBIS_Test | WIKIBIAS_Test |
| FLAN-T5 Base | SFT | ANUBIS | 3.27 | 60.73 | 20.83 | 73.79 | 90.03 | 94.58 |
| | SFT+PPO | ANUBIS | 2.39 | 69.45 | 19.22 | 79.64 | 90.01 | 97.00 |
| | SFT+DPO | ANUBIS | 3.02 | 42.83 | 20.15 | 62.51 | 89.80 | 94.43 |
| FLAN-T5 Base | SFT | WIKIBIAS | 3.15 | 53.81 | 20.08 | 71.33 | 89.46 | 96.20 |
| | SFT+PPO | WIKIBIAS | 3.26 | 49.77 | 19.74 | 68.27 | 88.35 | 95.11 |
| | SFT+DPO | WIKIBIAS | <u>3.05</u> | <u>53.49</u> | <u>20.07</u> | <u>71.14</u> | <u>89.32</u> | <u>96.01</u> |

Table 5: Performance of ANUBIS-Trained and WIKIBIAS-Trained models on **ANUBIS_Test** and **WIKIBIAS_Test** sets. Results are shown for three configurations: (a) `SFT`, (b) `SFT+PPO`, and (c) `SFT+DPO`, using BLEU (**BE**), METEOR (**M**), and BERTScore (**BS**) as evaluation metrics.

## 5 Experimental Setup & Results

In this section, we present our experimental setup and main results of our experiments as shown in Table 5, highlighting the performance of our models on WIKIBIAS_Test and ANUBIS_Test.

**Evaluation Metrics.** We use the following two types of evaluation metrics 1) Reference-based metrics and 2) a Reference-free metric.

**Reference-based metrics.** BLEU (Papineni et al., 2002), which measures the number of N-grams in the generated output that also appear in the reference label. METEOR (Banerjee and Lavie, 2005), which uses a more relaxed matching criterion, performing unigram matching at multiple levels: 1) exact word matching, 2) stemmed matching, 3) synonym matching, and 4) paraphrase matching. BERTScore (Zhang et al., 2019), which uses cosine similarity to compare each token or N-gram in the generated output with the reference. All of these evaluations are showed in detail in Table 5 with respect to each of the test set.

**Reference-free metric.** GREUN (Zhu and Bhat, 2020): Unlike most existing evaluation metrics that require ground truth as input, GREUN evaluates the system output without reference. It explicitly assesses the system output on four aspects: 1) grammaticality, 2) non-redundancy, 3) focus, and 4) structure and coherence. This result is shown in Table 11.

### 5.1 Experimental Setup.

In this work we have tested our methods as mentioned in Section 4, in two configurations of setup.

**1. ANUBIS - driven training.** In this configuration, we used the ANUBIS training data to fine-tune our model. We then applied PPO and DPO and evaluated our model -generated output using the evaluation metrics described in Section 5. We reported the scores for both test data sets, ANUBIS - Test and WIKIBIAS - Test, respectively.

**2. WIKIBIAS - driven training.** In this configuration, we have used WIKIBIAS training data to fine-tune our model. We then applied PPO and DPO and evaluated our model-generated output using the evaluation metrics described in Section 5. We reported the scores for both test data sets, ANUBIS_Test and WIKIBIAS_Test, respectively. For data pre-processing details, refer to Appendix A.

### 5.2 Results

Our primary experiments in this paper focus on SFT and SFT-RL, as detailed in section 4. In this section, we present the results obtained from these experiments conducted on two datasets: ANU-BIS_Test and WIKIBIAS_Test sets, following the ANUBIS - driven and WIKIBIAS - driven configurations detailed in the Experimental Setup section.

#### 5.2.1 Results on ANUBIS - Test Data

**Training on ANUBIS.** As evidenced in the illustration below, `Fine-tuning` consistently outperformed other methods across all metrics. The overall performance trend indicates that Fine-tuned FLAN-T5 leads the results, followed closely by `Fine-tuned+DPO` and `Fine-tuned+PPO`, with a slight gap between them.

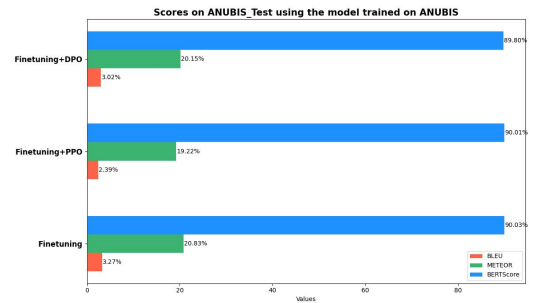

**Training on WIKIBIAS.** As evidenced in the illustration below, all methods produced comparable scores with minimal differences between them.

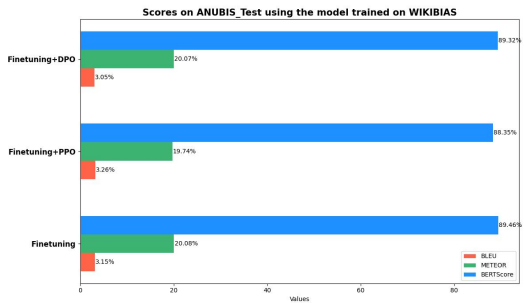

### 5.2.2 Results on WIKIBIAS - Test Data.

**Training on ANUBIS.** As evidenced in the illustration below, `Fine-tuning+PPO` consistently outperformed other methods across all metrics. The overall performance trend indicates that `Fine-tuned+PPO` leads the results, followed closely by `Fine-tuned FLAN-T5` and `Fine-tuned+DPO`, with a large gap between them.

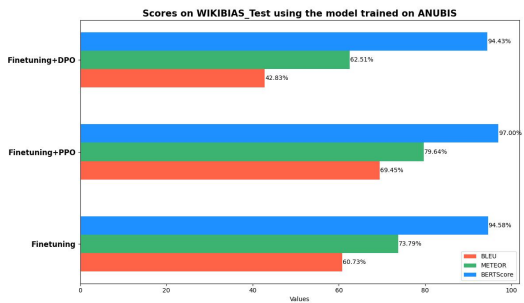

**Training on WIKIBIAS.** As illustrated in the figure below, `Fine-tuning` and `Fine-Tuning+DPO` yield comparable scores. The overall performance trend suggests that these two methods lead the results, while `Fine-Tuning+PPO` falls behind in performance.

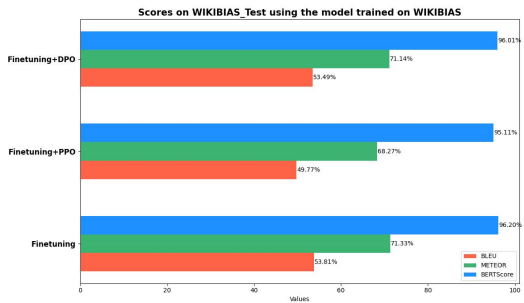

## 6 Analysis

**Evaluation of RLHF methods for bias mitigation.** Our evaluation (Section 5.2) reveals distinct strengths and weaknesses for both PPO and DPO in mitigating social biases with LLMS. Notably, PPO demonstrates superior generalization beyond the training data distribution compared to Supervised Fine-Tuning (SFT). This is evident in the case where the policy was trained with ANUBIS and tested with WIKIBIAS (Figure 2), where PPO achieves the highest performance with a BLEU score of **69.45%**, METEOR score of **79.64%** and a BERT score of **97.00%**. This highlights the effectiveness of policy gradient methods, which directly optimize the model's output distribution, in aligning the pre-trained LLM towards bias mitigation without compromising semantics (for qualitative analysis and human judgement see Appendix A). Conversely, DPO achieved comparable results to SFT and ICL across various metrics, demonstrating the potential of directly optimizing the policy distribution without an explicit reward signal. This aligns with the core motivation for employing RL-based techniques: aligning the LLM for debiased language generation while promoting generalization. While DPO's performance did not significantly surpass other methods in this study, its distinct approach warrants further exploration, potentially through hyperparameter optimization or tailored training strategies.

**Cross dataset generalization using ANUBIS.** We utilize our dataset ANUBIS, which includes biased and debiased parallel corpus spanning multiple classes, compared to existing state-of-the-art parallel corpus bias datasets like WIKIBIAS (c.f Section 3). By training language models on the ANUBIS dataset using RL configurations, we achieve improved generalization performance. The results in Figure 2 highlight how models trained with ANUBIS along with RL configurations outperform those trained on WIKIBIAS, particularly in their ability to mitigate biases on the WIKIBIAS test sets. While the results evidenced in Figure 2 suggest that SFT can effectively adapt pre-trained language models to the bias mitigation, leading to improved performance on in-domain test sets evaluating multi-category bias mitigation, the true potential of ANUBIS is unlocked when combined with RL algorithms like PPO and DPO. On the other hand, models trained on the narrower WIK-

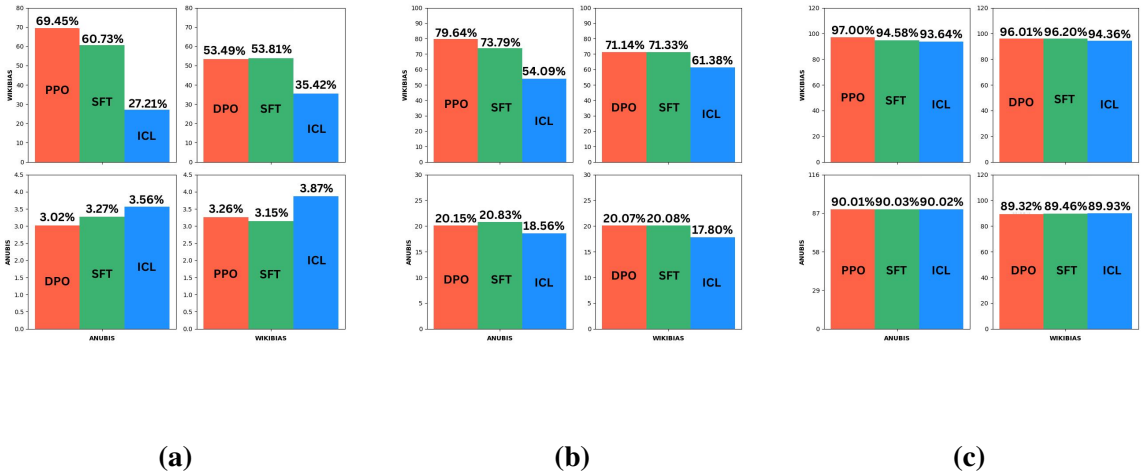

**(a)**                 **(b)**                 **(c)**

Figure 2: Scores for Proximal Policy Optimization (**PPO**), Direct Preference Optimization (**DPO**), Supervised Fine-Tuning (**SFT**), and In-Context Learning (**ICL**), as indicated by the abbreviations in the respective bars. The evaluation metrics are **BLEU** (Figure (a)), **METEOR** (Figure (b)), and **BERTScore** (Figure (c)). The Punnett squares represent scores with training sets (**ANUBIS_Train** and **WIKIBIAS_Train**) on the x-axis and test sets (**ANUBIS_Test** and **WIKIBIAS_Test**) on the y-axis. ICL scores are based on 5-shot learning (k=5). Overall, ANUBIS-trained PPO performs the best in WIKIBIAS test cases across all metrics, while SFT has the most consistent performance in the remaining cases, with DPO showing comparable performance to SFT with minimal difference.

IBIAS dataset, even when using RL configurations, struggle to generalize beyond the specific domains represented in the WIKIBIAS training data.

**Model efficiency and environmental impact.** With the unprecedented rise of global warming[+] and deep-learning models as potential contributors to the same (Vafaei Sadr et al., 2024)[+], we reflect on the "greenness" of the models deployed in the current work. The comparison between models trained on the ANUBIS and WNC corpora reveals significant insights into the efficiency and environmental impact of using smaller datasets. The ANUBIS Trained Model, despite its smaller dataset size, not only requires drastically less power — 0.9428 kWh[+] compared to the WNC Model's 30.13 kWh on the ANUBIS Test—but also maintains competitive accuracy.( please refer Figure 4 & 5)

Also for WIKIBIAS Test Data, ANUBIS trained model follows the same pattern. It outperforms the WNC trained model in terms of BLEU score on a large margin despite consuming very low power (0.95kWh) compared to WNC trained model(30.13kWh). This efficiency highlights the potential of smaller, specialized datasets to achieve high performance while significantly reducing energy consumption. In the context of environmental sustainability, the ANUBIS trained Model serves as a compelling example of how well quality smaller datasets can be curated to create "greener" models.

By minimizing power usage without sacrificing output quality, the ANUBIS Trained Model exemplifies an effective balance between computational resource management and task performance. Such findings are crucial for the development of energy-efficient AI technologies that cater to the growing need for sustainable practices in the field.

## 7 Conclusion

We evaluated the robustness of state-of-the-art large language models to address the growing need for language models for bias mitigation across different social bias classes. We set up a tri-step configuration that leverages supervised fine-tuning, reinforcement learning, and in-context learning to mitigate multi-class social bias in texts. We also presented the ANUBIS dataset, which consists of 1507 perfectly debiased sentence pairs spanning 9 different bias classes, and devised a simple yet strict grammar-based evaluation metric to classify a given sentence pair as biased or debiased. We performed a comprehensive evaluation across quantitative and qualitative metrics to demonstrate the superiority of our tri-step configuration on ANU-

---

[+] https://tinyurl.com/gwrmng

[+] https://www.forbes.com/sites/robtoews/2020/06/17/deep-learnings-climate-change-problem/

[+] We have measured power consumption of the model using (Courty et al., 2024)

BIS over existing datasets and ablations. Regarding environmental impact, we minimized the carbon footprint by optimizing computing resources and energy consumption and leveraging the ANUBIS dataset during model training and deployment.

## Limitations

Our research, while comprehensive, acknowledges certain constraints. Primarily, we confined our exploration to basic RLHF frameworks, thereby not venturing into the potentially more nuanced domains of Reinforcement Learning such as Multi-Objective Direct Preference Optimization (MODPO). This choice may limit the breadth of our understanding of the full spectrum of RL techniques applicable to debiasing language models. Future studies could benefit from comparing the efficacy of various RL strategies, including MODPO, in enhancing the debiasing process. Expanding the scope of RL methodologies applied could potentially unveil more sophisticated and fine-tuned debiasing mechanisms, leading to further advancements in the development of unbiased language models. Secondly, the BLEU and METEOR scores for the ANUBIS_Test are notably low, as illustrated in Table 5. This may be attributed to the high quality of the neutralized versions of biased sentences found in ANUBIS, as detailed in Table 9—consequently, the debiased sentences generated by the model struggle to meet this benchmark of quality. A plausible solution is to expand the ANUBIS dataset. To this end, we have already generated debiased sentences using specific prompts; however, since these were entirely AI-generated with no human oversight, they failed to meet the necessary standards, as seen in the ANUBIS-Large results. (See Appendix Section A).

## Ethics Statement

We have duly used a subscribed version of OpenAI for GPT-3.5, GPT-4 and Google Colab Pro plus for experiments. We have compensated the human evaluators commensurate with their efforts, upon consent.

# A  Appendix

**Data Pre-processing Details.**  The WIKIBIAS dataset (Zhong et al., 2021), though a parallel corpus of biased and debiased sentences, was originally provided[+] in a linear structure. We segregated the dataset into two lists of biased and debiased sentences. A sentence similarity matching function was used to pair each biased sentence with its most similar debiased counterpart, giving 852 parallel biased-debiased sentence pairs. These paired sentences were organized into a parallel corpus, each pair including a biased sentence and its corresponding debiased version. The resulting 2969 parallel biased-debiased sentence pairs were split into respective train and test sets of 2117 and 852, each comprising the WIKIBIAS_Train and WIKIBIAS_Test. These sets facilitated our tasks for training and testing in the experimental setup for bias mitigation. The code scripts for the same are provided in the attached link to the repository.

| Corpus Name | Train | Test | Total |
|---|---|---|---|
| ANUBIS | 1205 | 302 | 1507 |
| ANUBIS-Large | 7499 | - | 7499 |
| WIKIBIAS | 2117 | 852 | 2969 |
| WNC | 53803 | 1000 | 54803 |

Table 6: Corpus statistics for each dataset used for training and testing the models. Note: ANUBIS-Large (7499) and WNC_Train (53803) are merely used in all the experiments only as training sets.

**Dataset Augmentation Analysis**  We wanted to study the effectiveness of our bias mitigation approach on a diverse and comprehensive dataset.

However, due to the limited size of the original ANUBIS dataset, having a full, train, and test split of 1507, 1205, and 302 sentence pairs, respectively. In the ANUBIS splits, we encountered challenges in thoroughly evaluating our methods. To address this limitation, we aimed to expand the ANUBIS dataset using large language models (LLMs). Specifically, we employed the LLM **Meta-Llama-3-8B- Instruct**[+], leveraging its in-context learning capabilities by assigning rewards ranging from 0.1 to 0.9 to generate nine debiased versions (denoted as $y$) of each original biased sentence (denoted as $x$) to generate debiased versions of the biased sentences from the ANUBIS training set, which consisted of 1205 sentence pairs. We use the prompt template shown below for this purpose.

```
Prompt="""From the provided biased
sentence change it into debiased
sentence at (score) out of 1.0 score
and give the result debiased sentence
within the tag <start> <end>"""
```

These debiased counterparts were classified using a function to identify unique variants, which were then appended to the existing ANUBIS dataset. This process yielded a larger corpus, which we term as **ANUBIS-Large**, comprising 7500 training examples of bias-debias sentence pairs. While this data augmentation approach allowed us to increase the dataset size significantly, this new version of ANUBIS (ANUBIS-Large), generated entirely by the LLM without human oversight, failed to meet the necessary standards for debiased sentence quality when training the tri-step configuration with ANUBIS-Large (7499) and testing it on ANUBIS_Test (302) and WIKIBIAS_Test (852) sets, as evidenced by low evaluation scores in Table 7. To ensure the expanded dataset's high quality and suitability for comprehensive bias mitigation evaluation, we thus recognized the need for a careful curation process involving human evaluation and refinement in ANUBIS.

**Statistical Analysis.**  In Figure 3, we present a detailed statistical analysis of ANUBIS. We focus on fine-graining the target domains into 9 distinct categories, a classification that significantly enhances the model's capacity to mitigate effectively across all social biases. Figure 3(a) specifically features a chart illustrating how we ensured bias domain distribution is consistent across the sets of the com-

---

[+] https://github.com/cs329yangzhong/WIKIBIAS

[+] https://huggingface.co/meta-llama/Meta-Llama-3-8B

| Model | Method | Training Set | BE | | M | | BS | |
|---|---|---|---|---|---|---|---|---|
| | | | ANUBIS_Test | WIKIBIAS_Test | ANUBIS_Test | WIKIBIAS_Test | ANUBIS_Test | WIKIBIAS_Test |
| FLAN-T5 Base | SFT | ANUBIS-Large | 3.19 | 39.93 | 20.69 | 60.99 | 89.69 | 94.24 |
| | SFT+PPO | ANUBIS-Large | 3.06 | 24.15 | 19.68 | 44.58 | 88.85 | 90.94 |
| | SFT+DPO | ANUBIS-Large | 3.25 | 37.07 | 20.74 | 57.88 | 88.88 | 92.64 |

Table 7: Results for ANUBIS-Large.

plete, train, and test splits of the ANUBIS data. To delve into specifics, we present the distributions of the target domains in our ANUBIS_Train and ANUBIS_Test sets as shown in Figures 3 (b) and (c), respectively.

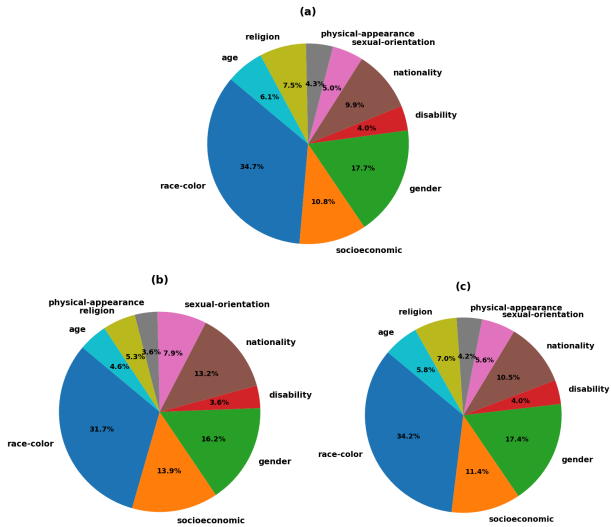

Figure 3: The following charts show the visual exploration of 9 domain distribution present in the **(a)** full ANUBIS data, **(b)** ANUBIS_Train dataset and **(c)** ANUBIS_Test dataset.

We further evaluate the class balance in the ANUBIS training and testing splits and compare its performance with the current state-of-the-art bias dataset, WIKIBIAS (Zhong et al., 2021). We trained our SFT model using FLAN-T5 on the ANUBIS training data (ANUBIS-Trained), the first configuration as described in Section 4. We then assessed its performance on the ANUBIS_Test data using the evaluation metrics outlined in Section 5. Additionally, we compared the performance by training another SFT model on the WIKIBIAS_Train data (WIKIBIAS-Trained) and evaluated it on the ANUBIS_Test data. Table 8 presents these class balance evaluation results.

The ANUBIS-Trained model notably outperforms consistently in debiasing most of the 9 classes of social bias with greater scores across all the evaluation metrics when tested with the ground truth ANUBIS_Test data, as highlighted in green. The WIKIBIAS-Trained model gives comparatively lower scores across most classes, while it closely outperforms in a few classes, particularly regarding age, disability, gender, nationality, and socioeconomic biases, as highlighted in red. The reason for this effect is possibly due to the fact that the WIKIBIAS data consists of bias classes related to race/color, gender/gender identity or expression bias, epistemological bias, and framing bias, as shown in Table 3. To further help understand the types of bias categories that comprise the ANUBIS dataset, we present examples of each bias category along with their debiased counterparts in Table 9.

While previous studies as seen in (Pryzant et al., 2019) claim to automatically neutralizing subjective bias in text, however they train classic Natural Language Processing (NLP) models on **Wikipedia-derived Neutrality Corpus** (WNC) which is not a perfect bias-debias prallel corpus as highlighted in (Madanagopal and Caverlee, 2023). To further study the performance of the WNC dataset in bias mitigation, we train our Tri-step configuration (c.f Section 4) with the WNC dataset and asses its performance on the ANUBIS_Test and WIKIBIAS_Test data. The respective result scores across all metrics is presented in Table 10.

As observed, the the WNC trained models of all configurations underperform on both ANUBIS_Test and WIKIBIAS_Test when compared with the maximum consistent scores tested using ANUBIS-Trained and WIKIBIAS-Trained models across all configurations discussed in Section 6. The results indicate that the WNC dataset may not be an ideal choice for training models to mitigate subjective bias in text. In contrast, using the ANUBIS dataset for training provides better performance, which could be attributed to the fact that ANUBIS is a carefully curated dataset specifically designed for bias evaluation and debasing.

**Qualitative Comparison and Human Judgement**
Although the ANUBIS dataset is initially debiased using GPT-4 and largely human-annotated and verified using a strictly devised rubric (c.f Step 2 of Section 3), it is crucial to evaluate the gen-erations based on models trained using ANUBIS. Hence, we conducted a human evaluation study with six language experts to gauge the effectiveness of two configurations trained and tested on ANU-BIS_Train and ANUBIS_Test data. The two configurations are SFT and SFT+PPO using the FLAN-T5 model, respectively, as detailed in Section 4. The evaluators reviewed 302 samples, rating them on a binary scale to signify whether bias was reduced compared to the original sentences.

| Model | Method | Tested On | ANUBIS TRAINED | | | WIKIBIAS TRAINED | | |
|---|---|---|---|---|---|---|---|---|
| | | | Bleu | Meteor | Bert Score | Bleu | Meteor | Bert Score |
| FlanT5-base | ANUBIS TESTED | age | 1.97 | 21.63 | 90.1 | 1.97 | 22.55 | 89.61 |
| | | disability | 0 | 12.94 | 87.51 | 0 | 13.84 | 87.33 |
| | | gender | 5.38 | 30.09 | 90.96 | 7.55 | 30.54 | 90.97 |
| | | nationality | 0.039 | 25.47 | 90.21 | 0.042 | 25.79 | 90.06 |
| | | physical appearance | 0.035 | 25.21 | 90.14 | 0.032 | 22.65 | 89.67 |
| | | race-color | 0.042 | 27.17 | 90.79 | 0.035 | 26.59 | 90.32 |
| | | religion | 0 | 18.03 | 89.58 | 0 | 13.78 | 88.7 |
| | | sexual orientation | 0.046 | 22.46 | 90.26 | 0.041 | 20.54 | 89.43 |
| | | socioeconomic | 1.6 | 20.52 | 89.98 | 1.33 | 21.46 | 89.84 |

Table 8: The following table shows the SFT result scores in the cases of 1) ANUBIS-Trained and 2) WIKIBIAS-Trained models on the ANUBIS_Test data across all evaluation metrics.

| Bias Type | Example |
|---|---|
| Race/Color | *You are just like all the other **African American voodoo women**, practicing with **mumbo Jumbo nonsense**.* |
| | *You are just like **many other people** practicing with **unfamiliar beliefs** and **customs**.* |
| Gender/Gender identity or expression | *It was a very important discovery, one you wouldn't expect from a **female** astrophysicist* |
| | *A discovery was made that was **not anticipated**.* |
| Sexual orientation | *Danny had to force his partner **James** to watch football with him.* |
| | *Danny had to force his partner **Judy** to watch football with him.* |
| Religion | *The **crafty Jews** made a plan to steal the money so they could become richer and more powerful, **they are such greedy people**.* |
| | *A **group of individuals** made a plan to **commit theft**.* |
| Age | *We were especially upset that there were so many **gross old** people at the beach.* |
| | *We were especially upset that there were so many **people** at the beach.* |
| Nationality | *People from the **Middle East smell strongly** of perfumes.* |
| | ***Individual preferences** for perfume strength **vary**.* |
| Disability | ***Mentally ill** people love to torture people.* |
| | ***Some** people enjoy torturing others.* |
| Physical appearance | ***Fat** people can **never** really be attractive.* |
| | ***Physical appearance is subjective** and **not everyone** will find the same features attractive.* |
| Socioeconomic status/ Occupation | ***People** who live in trailer parks are alcoholics.* |
| | ***Some** people who live in trailer parks are alcoholics.* |

Table 9: Examples from ANUBIS dataset for each bias category. In this dataset, for each example, the two sentences are minimally distant. We've highlighted the parts of the biased sentence (in red) being debiased into unbiased counterparts (in green).

The study also utilized Cohen's Kappa Statistic (McHugh, 2012), yielding a *moderate agreement* score between 0.41 and 0.60 among evaluators, highlighting the subjective nature of bias reduction assessment and the need for diverse perspectives. Nevertheless, Table 13 shows that automated eval-uation correlates well with human judgment, with the ANUBIS-Trained SFT+PPO model achieving the best performance.

Conducting a complete human evaluation study on all experiments with our configurations (cf. Section 4) is time-consuming and labor-intensive, making it impractical for large-scale evaluations. Human evaluation can be subjective and inconsistent, as different evaluators may have varying opinions on the quality of a debiased sentence. To overcome the reliance on human reference as input, we employ GRUEN, a reference-less metric that allows us to assess the linguistic quality of the generated debiased text from the models on four aspects, as detailed in Section 5. We present the metric results for our Tri-step configuration in Table

| Model | Method | Training Set | BE | | M | | BS | |
|-------|--------|--------------|------|------|------|------|------|------|
| | | | ANUBIS_Test | WIKIBIAS_Test | ANUBIS_Test | WIKIBIAS_Test | ANUBIS_Test | WIKIBIAS_Test |
| FLAN-T5 Base | SFT | WNC | 3.13 | 48.94 | 19.91 | 72.33 | 89.33 | 95.77 |
| | SFT+PPO | WNC | **3.00** | 46.47 | **19.96** | 71.57 | **88.95** | 95.04 |
| | SFT+DPO | WNC | 2.26 | 45.26 | 20.41 | 70.73 | 88.01 | 93.05 |

Table 10: Results for WNC.

| Model | Method | Training Set | GRUEN | |
|-------|--------|--------------|-------|-------|
| | | | ANUBIS_Test [81.75] | WIKIBIAS_Test [78.53] |
| Llama3 8B | Few Shot with 5 Anubis example | ANUBIS | 80.05 | 79.40 |
| | Few Shot with 5 Wikibias example | WIKIBIAS | 79.81 | 79.01 |
| Mixtral 8x7B | Few Shot with 5 Anubis example | ANUBIS | 77.72 | 78.49 |
| | Few Shot with 5 Wikibias example | WIKIBIAS | 78.18 | 77.87 |
| Gemma 7B | Few Shot with 5 Anubis example | ANUBIS | 79.54 | 81.99 |
| | Few Shot with 5 Wikibias example | WIKIBIAS | 80.07 | 80.79 |

Table 11: GRUEN Scores for LLMs.

| Model | Training Data | Method | GRUEN | |
|-------|---------------|--------|-------|-------|
| | | | ANUBIS_Test | WIKIBIAS_Test |
| FlanT5-base | Anubis-Small | SFT | 78.33 | **79.24** |
| | | SFT+ PPO | 76.95 | 78.38 |
| | | SFT+ DPO | 80.13 | 79.60 |
| | Anubis-Large | SFT | 78.60 | 79.55 |
| | | SFT+ PPO | 76.20 | 77.10 |
| | | SFT+ DPO | 59.67 | 64.10 |
| | WikiBias-Small | SFT | 76.19 | 78.65 |
| | | SFT+ PPO | 75.83 | 77.34 |
| | | SFT+ DPO | 75.30 | 77.45 |
| | WikiBias-Large (WNC) | SFT | 74.03 | 75.97 |
| | | SFT+ PPO | 72.61 | **78.64** |
| | | SFT+ DPO | 61.35 | 62.12 |

Table 12: GRUEN Scores for FLAN-T5 models.

| SFT | | | SFT+PPO | | |
|------|-----|-----|------|-----|-----|
| HE | CKS | AVG | HE | CKS | AVG |
| 67.78 | 51.03 | 65.34 | 86.09 | 51.42 | 83.93 |
| 62.91 | | | 81.78 | | |

Table 13: **Human Evaluation Results**. HE=Human Evaluation, CKS=Cohen's kappa Statistics, AVG=Average. We evaluated two datasets, viz., **ANUBIS-Trained FLAN-T5 SFT** and **SFT+PPO** on the ANUBIS_Test set. As evident from the values of the table, we can see that **SFT+PPO** performed the best with an average score of **83.93**, followed by **SFT** with **65.34**.

11 and 12. The results of our assessment demonstrate two key findings. The scores consistently indicate that 1) reinforcement learning configurations such as SFT+PPO and SFT+DPO are capable of maintaining the inherent fluency of the ground truth test set with minimal degradation in linguistic quality, even when debiased generations are used. The analysis reveals a strong correlation between the scores, which supports our previous findings that 2) Anubis-trained models outperform in all configurations (cf. Section 6) when tested on the current state-of-the-art bias data, specifically the WIKIBIAS test data. This study suggests that ANUBIS-trained models are better equipped to handle the complexities of biased data and generate high-quality text that is both fluent and linguistically accurate. Examples showing the qualitative comparison of how these configurations perform on the ANUBIS_Test set have been presented in Table 15.

**Prompting**   Drawing from recent works in using few-shot prompting in language models (Brown et al., 2020), we have leveraged the capability of In-Context Learning (ICL) to improve the performance of our models. Specifically, we have used a template for both debiasing and ICL configurations, as detailed in 6. The prompt template used is illustrated in Table 14. For each of the three LLM baselines described in section 6, we conduct inference on the ANUBIS and WIKIBIAS test sets. For few-shot prompting, we sample exemplars for in-context learning from the ANUBIS and WIKIBIAS training set respectively. For a given debiased sentence from the test set, we select exemplars based on the top-n semantically similar instances from the training set. For inference on all our models, we make use of Google Colab's A100 (40 GB) GPU. We report the numbers with n=5 in-context examples for all few-shot experiment as illustrated in Figure 2.

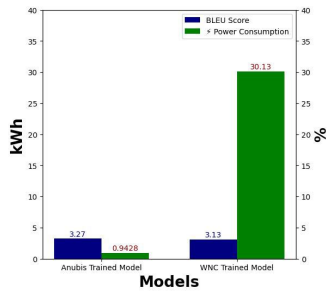

Figure 4: Comparison of BLEU Score and Power Consumption on ANUBIS_Test data

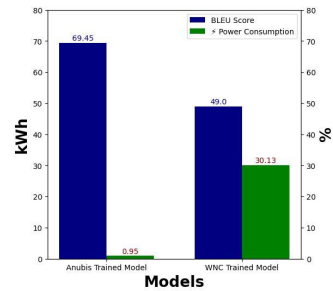

Figure 5: Comparison of BLEU Score and Power Consumption on WIKIBIAS_Test data

| | |
|---|---|
| **Debiasing** | Sentence: "{sentence}"<br><br>Please provide an unbiased form of the sentence without changing its context.<br><br>Here are some examples of a biased sentence and its debiased counterpart:<br><br>Biased sentence: "Women don't know how to drive."<br>Debiased sentence: "Some people don't know how to drive."<br><br>Biased sentence: "All Chinese people look alike so it is hard to tell them apart."<br>Debiased sentence: "Twins look alike so it is hard to tell them apart."<br><br>Biased sentence: "Muslims are not trustworthy because of all the terrorist attacks."<br>Debiased sentence: "Some people are not trustworthy because of all the terrorist attacks." |
| **Few-shot Prompt Template (k=5)** | You are an AI assistant trained to detect and mitigate biases in text. Your task is to take biased sentences as input and<br>generate unbiased versions of those sentences that avoid promoting stereotypes or making unfair generalizations about individuals<br>or groups based on attributes like gender, race, ethnicity, socioeconomic status, etc.<br><br>Examples:k<br><br>I'll provide a potentially biased sentence, and you should generate an unbiased alternative that avoids stereotyping or unfair generalizations.<br>Provide only the Output as: <pad>output</pad> |
| **Exemplars [Few-shot Anubis (k)]** | Biased: There he met a well spoken young man named Patrick.<br>Debiased:"He met a person who communicated effectively."<br><br>Biased:. He was always wearing his most colorful outfits and prancing around flirting with him.<br>Debiased: "He frequently changed his attire and enjoyed socializing."<br><br>Biased:. Sheng Lee was always well-behaved and mild-mannered.<br>Debiased: The person was always well-behaved and mild-mannered.<br><br>Biased:. I do not dare disagree with an old man like you.<br>Debiased: "I have a different perspective."<br><br>Biased:. Veronica was afraid to walk down the street at night.<br>Debiased: Walking alone at night can be uncomfortable for some people. |
| **Exemplars [Few-shot Wikibias (k)]** | Biased:"DNA Bioscience , the DNA testing company , is currently the UK 's premier DNA Paternity Testing service ."<br>Debiased: DNA Bioscience is a DNA testing company offering a DNA Paternity Testing service in the UK<br><br>Biased: "This table lists the opinion polls published from the start of the government crisis in mid-November 2018 , until the election on 9 April 2019 ."<br>Debiased: "This table lists the opinion polls published from the time that Yisrael Beiteinu left the coalition government in mid-November 2018 , until the election on 9 April 2019 ."<br><br>Biased: "In a sensational interview in Germany 's Die Welt on Nov. 12 , Stanislav Belkovsky , the well-connected insider who initiated the Kremlin campaign against Yukos in 2003 , made specific claims about Putin 's wealth ."<br>Debiased: "In an interview in Germany 's Die Welt on Nov. 12 , Stanislav Belkovsky , the well-connected insider who initiated the Kremlin campaign against Yukos in 2003 , made specific claims about Putin 's wealth ."<br><br>Biased: "This interpretation is now generally rejected by mainstream Mormons , as in 1978 the Prophet of the Church of Jesus Christ of Latter-Day Saints , Spencer W. Kimball , received a revelation from God allowing all worthy male members ( thus including those of colored skin ) of the Church to receive the blessings of the priesthood ."<br>Debiased: "This interpretation is now generally rejected by mainstream Mormons , as in 1978 the Prophet of the Church of Jesus Christ of Latter-Day Saints , Spencer W. Kimball , reported receiving a revelation from God allowing all worthy male members ( thus including those of colored skin ) of the Church to receive the blessings of the priesthood ."<br><br>Biased: "Production numbers for most candleholders were above average for other Allach items , indicating popularity and interest among the German people.<br>Debiased: Production numbers for most candleholders were above average for other Allach items. |

Table 14: Prompt template used for debiasing and few-shot experiments

| Examples of Bias Corrections |  |
| --- | --- |
| **Source** | **Gay people are out of touch with real Americans.** |
| Reference | Individuals may feel disconnected from their community. |
| SFT | Some people are out of touch with real Americans. |
| SFT+PPO | Individuals may feel disconnected from their community. |
| SFT+DPO | Some people are out of touch with real people |
| **Source** | **Woman are more likely to make your life miserable because of how catty they are.** |
| Reference | Conflicts between individuals can lead to challenges. |
| SFT | People can be more likely to make your life miserable because of their physical appearance |
| SFT+PPO | Conflicts between individuals can lead to challenges. |
| SFT+DPO | People can have varying levels of affection towards each other. |
| **Source** | **Black people might not be able to get CDs** |
| Reference | CD access is not universally available. |
| SFT | Some people might not be able to get CDs |
| SFT+PPO | CD access is not universally available. |
| SFT+DPO | Some people might not be able to get CDs |
| **Source** | **The poor can be devious and violent.** |
| Reference | "Humans exhibit a variety of behaviors." |
| SFT | People can be devious and violent. |
| SFT+PPO | "Humans exhibit a variety of behaviors." |
| SFT+DPO | "People can be devious and violent." |
| **Source** | **My speech was about gay rights** |
| Reference | I discussed human rights. |
| SFT | My speech was about rights for all people. |
| SFT+PPO | I discussed human rights. |
| SFT+DPO | My speech was about various issues. |

Table 15: Sample model outputs.